%% file: main.tex
\documentclass{llncs}
\usepackage[utf8]{inputenc}
\usepackage[misc]{ifsym}

\usepackage{amsmath}
\usepackage{graphicx}
\usepackage{hyperref}
\usepackage{placeins}
\usepackage{subcaption}
\usepackage{fancyhdr}
\usepackage{color}
\usepackage{soul}

\begin{document}
	\title{Evolutionary Construction of Convolutional Neural Networks} 
    \author{Marijn van Knippenberg\inst{1}\textsuperscript{(\Letter )}\and Vlado Menkovski\inst{1}\and Sergio Consoli\inst{2}}
    \institute{Eindhoven University of Technology (TU/e)\\\email{\{m.s.v.knippenberg,v.menkovski\}@tue.nl} \and Philips Research Eindhoven\\\email{sergio.consoli@philips.com}}
    
    \maketitle
    
    \begin{abstract}
    	\input{sections/0_abstract}
    \end{abstract}
    
    
    \input{sections/1_introduction}
    
    \input{sections/2_related}

    \input{sections/3_method}

    \input{sections/4_experiments_results}
    
    \input{sections/5_conclusion}
        
    \bibliographystyle{splncs03}
    \bibliography{thesis.bib}
\end{document}

%% file: sections/0_abstract.tex
Neuro-Evolution is a field of study that has recently gained significantly increased traction in the deep learning community. 
It combines deep neural networks and evolutionary algorithms to improve and/or automate the construction of neural networks. Recent Neuro-Evolution approaches have shown promising results, rivaling hand-crafted neural networks in terms of accuracy.

A two-step approach is introduced where a convolutional autoencoder is created that efficiently compresses the input data in the first step, and a convolutional neural network is created to classify the compressed data in the second step. The creation of networks in both steps is guided by by an evolutionary process, where new networks are constantly being generated by mutating members of a collection of existing networks. Additionally, a method is introduced that considers the trade-off between compression and information loss of different convolutional autoencoders. This is used to select the optimal convolutional autoencoder from among those evolved to compress the data for the second step.

The complete framework is implemented, tested on the popular CIFAR-10 data set, and the results are discussed. Finally, a number of possible directions for future work with this particular framework in mind are considered, including opportunities to improve its efficiency and its application in particular areas.

%% file: sections/1_introduction.tex
\section{Introduction}

Neural networks have become a popular data analysis tool in both academia and industry, especially when tasks like image classification, natural language processing, and speech recognition need to be addressed. They have shown to be very adept at these tasks, which are classic problems in which we like the computer to show ``human-like'' behavior. One of the challenges surrounding neural networks which has kept researchers around the world occupied is the matter of their design, and the possible automation of this task. Constructing a neural network is still often seen as a somewhat ``magic'' skill by many: a combination of knowledge, past experience, and intuition. Network performance is greatly affected by its size and structure, the type and order of layers, the choice of loss function and the way data is presented to the network. These are all decisions that a designer has to make in what is often a lengthy continuous cycle of re-design, re-training, and re-evaluation. This makes it an attractive target for automation \cite{DBLP:journals/corr/MiikkulainenLMR17}. 

In this paper we introduce a neuro-evolution approach to the above challenge that is more efficient than current solutions in terms of computational resource consumption \cite{DBLP:journals/corr/MiikkulainenLMR17}. This will hopefully make this type of approach more viable in real-life settings. It takes inspiration from a specific, recently published study, which establishes a kind of baseline for the application of evolutionary algorithms in this setting \cite{DBLP:conf/icml/RealMSSSTLK17}. Implementation of the approach also requires a look at solution efficiency and multi-criteria decision making (MCDM), which helps reasoning about the relation between the performance metrics of different types of neural networks \cite{DBLP:journals/corr/abs-1712-06567,DBLP:conf/gecco/MorseS16}.

Both the baseline and the proposed approach are implemented and tested in order to confirm their behavior and to draw comparisons. The results are then inspected to establish whether the new framework has significant impact in terms of final accuracy and consumed computational resources. These two will form a trade-off that is controlled by the MCDM process. Finally, a number of alternatives and future study topics are presented and considered.

%% file: sections/2_related.tex
\section{Related Work}

Neuro-evolution is the field of study that combines neural networks with evolutionary algorithms in the search for innovative training methods \cite{DBLP:journals/evi/TurnerM14}. The field has recently gained increased interest with the sharp increase of popularity of deep neural networks. Likely the most well-known algorithm in the field of neuro-evolution is the neuro-evolution of augmented topologies algorithm (NEAT) \cite{stanley2002evolving}. Recent studies have applied evolutionary strategies to deep neural networks in a variety of manners, such as evolving network weights via genetic programming \cite{DBLP:conf/gecco/MorseS16,DBLP:conf/gecco/SuganumaSN17}, differential evolution \cite{DBLP:journals/cin/RereFA16}, pitting networks against each other in a tournament selection environment \cite{DBLP:conf/icml/RealMSSSTLK17}, more recent extensions of NEAT \cite{DBLP:conf/gecco/Desell17,DBLP:journals/corr/MiikkulainenLMR17}, and by evolving a network’s activation functions \cite{DBLP:conf/gecco/HaggMA17}. The general ``mood'' of the intersection of these two fields is very much an exploratory one. There is no clear precedent on how to apply techniques from the meta-heuristics field to deep learning \cite{DBLP:conf/gecco/HaggMA17}. Another interesting direction in neuro-evolution that has recently been taken under the growing complexity of deep neural networks is that of network reduction. Evolutionary strategies can be used to simplify networks, in the hope of reducing future training and operating cost \cite{DBLP:journals/corr/ShafieeBW17}. 

This work is based in large parts on a recent Neuro-Evolution study \cite{DBLP:conf/icml/RealMSSSTLK17}. The goal of this study was to establish the viability of genetic programming in replacing human input during the design phase of neural networks. In the remainder of this paper, their approach will be referred to as the ``baseline framework''. The approach is based on multiple worker processes operating in parallel, each applying a genetic programming algorithm that continually evolves a population of neural networks. In genetic programming, a collection of candidate solutions, called the \emph{population}, is updated in steps in the hopes of finding better candidate solutions. A candidate solution is also known as an \emph{individual}. An individual is considered to have ``DNA''; some simplified representation from which the solution can be reconstructed. By mutating this DNA in different ways, an individual can be evolved into new individual, one which hopefully has better \emph{fitness}, which is some measure of the quality of the solution. In this case, each candidate solution is a convolutional neural network, and the DNA is simply the structure of the network. In order to apply genetic programming, a \emph{selection method} is required. This is the method that is used to determine which individuals in the population are chosen for mutation. The approach is based on the basic process of tournament selection \cite{DBLP:journals/compsys/MillerG95}. It is a simple and direct selection method that is based on ``rounds''. Each round, $k$ individuals are selected from the population according to some probability distribution, and their fitness is compared. The individual with the best fitness is copied and its copy is mutated before inserting it into the population, while the one with the worst fitness is removed from the population. This method of selecting individuals is intuitive, easy to implement, and easy to adjust. An added benefit is that it is quite easy to parallelize. In the case of the study, $k=2$ 
individuals are selected uniformly at random. In order to establish a population to select from, each worker process generates some initial individuals. In this case, the initial individuals are very simple networks with very low fitness scores. 

The major issue of this approach is the computational resource required. To arrive at a competitive result, many networks have to be generated, trained, and evaluated. The authors' main goal was to establish viability in terms of accuracy, so their approach is fairly basic and direct. This leaves many opportunities for improvements and further study, which is the point where this work steps in.

%% file: sections/3_method.tex
\section{Method}

\subsection{Evolving Autoencoders}

The attempt of this work to reduce computation time is based on reducing the size of the input samples that are used for training classification networks. Reducing the input sample size has a double positive effect on the evolutionary process: training each individual network takes less effort, since there are less values to process for each sample, and networks can be expected to be shallower since the input is smaller, reducing the average training time per network in the overall evolutionary process. 

Autoencoders are well-suited tools for reducing sample sizes, and in the case of image data, a convolutional autoencoder can be used to maintain the spatial relations of the input data in the encoded data. Since convolutional autoencoders are simply convolutional neural networks with some extra restraints and a different error metric, the baseline framework can also be applied to obtain the most suitable autoencoder. The overall process is then as follows:

\begin{enumerate}
	\item Perform an evolutionary process on a population of convolutional autoencoders. Training is based on the original input and the reconstruction error of the decoder.
    \item Pick the best autoencoder and encode the entire original input data set.
    \item Perform an evolutionary process on a population of convolutional neural networks that classify input samples. Training is based on the encoded input and classification error.
    \item Pick the classifier with the highest validation accuracy and append it to the encoder used to encode the data in order to obtain the best overall network.
\end{enumerate}

In order to apply the baseline framework to convolutional autoencoders, a new selection process needs to be refined. In the baseline framework, selection was purely based on accuracy. In the case of autoencoders, there are now two selection criteria: reconstruction accuracy and compression ratio. We like both of these to be as high as possible, but it is important to consider the case where one autoencoder has a higher compression ratio, while another one has a higher accuracy. Taking inspiration from the Non-dominated Sorting Genetic Algorithm (NSGA)\cite{DBLP:journals/tec/DebAPM02}, autoencoders are grouped into Pareto fronts based on dominance. If two sampled individuals are members of different fronts, the one that is a member of the dominating front ``wins''. If both individuals are part of the same front, neither is strictly better than the other in terms of Pareto efficiency. In this case, the individual that is furthest away the other individuals in that front is chosen as winner. In other words, a higher probability of innovation is rewarded if both individuals come from the same front. The final adjustment to make to the evolutionary process is to define a new set of mutations so actual convolutional autoencoders are evolved.

\begin{figure}[ht]
	\centering
	\includegraphics[width=0.75\textwidth]{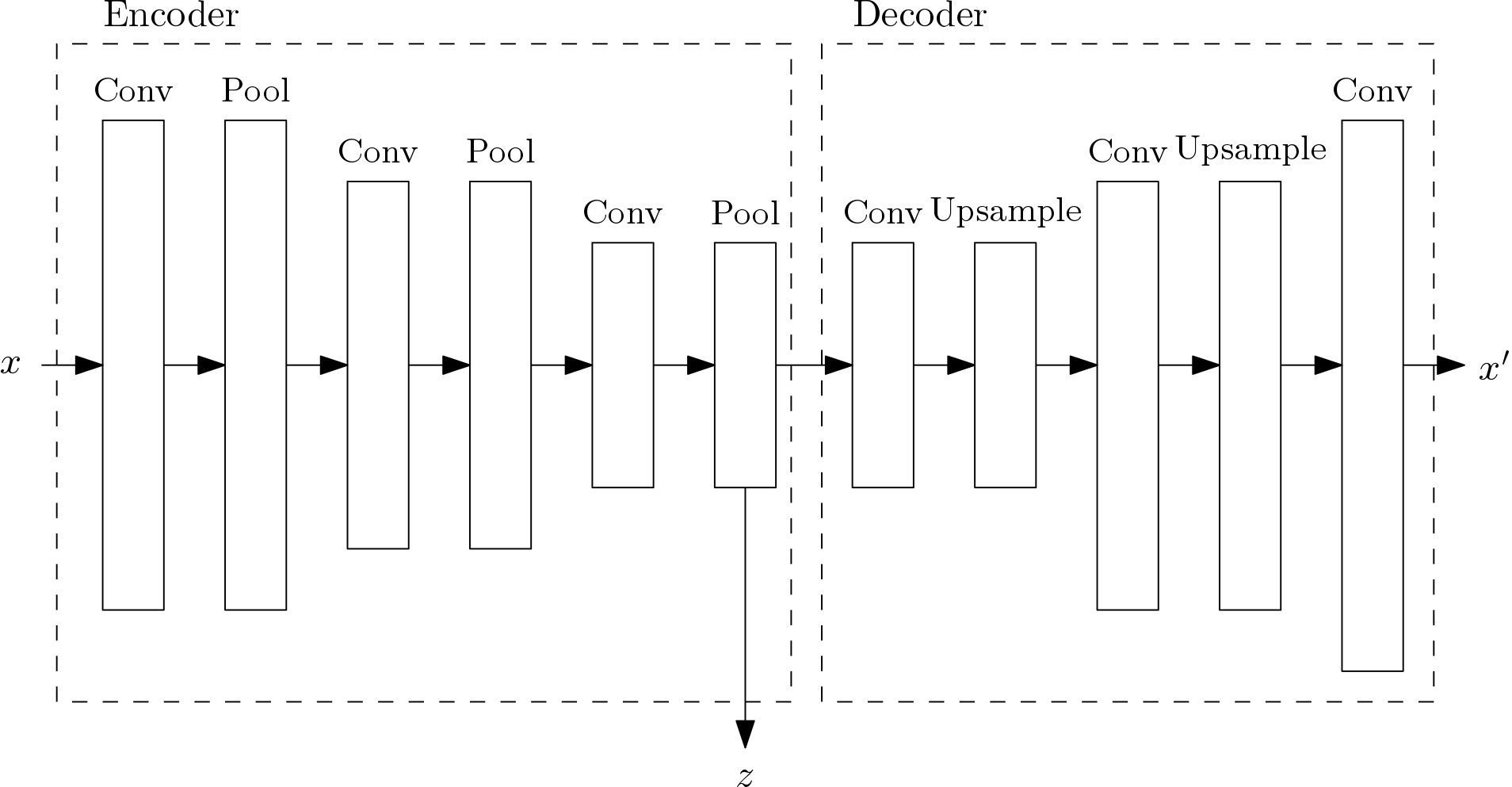}
	\caption{Convolutional autoencoder structure. For some input $x$, the encoder produces an encoding $z$, and the decoder produces a reconstruction $x'$. The decoder mirrors the encoder, replacing pooling layers with up-sampling layers.}
    \label{fig:cae}
\end{figure}

Below are listed the mutations that are made available to the evolutionary framework for the purpose of evolving convolutional autoencoders. These are based on the baseline framework, with some small changes, such as the inclusion of pooling-related mutations. Note that only the encoder is mutated, since the decoder is directly derived from it by mirroring it and replacing each pooling layer with an up-sampling layer (see Figure \ref{fig:cae}). 

\begin{itemize}
	\item \textsc{Identity}: The network structure is not changed in any way. In practice this means the same network trained longer, since weights are maintained when copying networks.
    \item \textsc{Insert convolution}: This mutation inserts a convolutional layer at a random location in the network in the encoder.
    \item \textsc{Remove convolution}: This mutation removes a random convolutional layer from the encoder.
    \item \textsc{Alter Stride}: The stride length of a random convolutional layer is incremented or decremented by 1 at random.
    \item \textsc{Insert pool}: This mutation inserts a max-pooling layer at a random location in the encoder. The initial pool size is 2x2.
    \item \textsc{Remove pool}: In this case a random max-pooling layer from the encoder is removed.
    \item \textsc{Alter filter number}: The number of filters of a random convolutional layer is adjusted.
    \item \textsc{Alter filter size}: One of the dimensions of the filters of a random convolutional layer is incremented or decremented at random
    \item \textsc{Alter pool size}: The pool size of a random max-pooling layer is incremented or decremented at random.
\end{itemize}

On top of the mutations themselves, there is an additional constraint on each mutation. For a mutation to be applied successfully, the output size of the encoder must be smaller than input size of the encoder. This forces the evolutionary process to evolve the encoder that actually compresses the data, and helps preventing the autoencoder from learning a dictionary of the input. 

At the end of the evolutionary process, one of the autoencoders has to be chosen from the Pareto front to process the original data set to obtain a compressed data set that is used in the second evolutionary process. A method is needed that considers the trade-off between the compression ratio and the accuracy of the autoencoders. This relies on 
MCDM, since the best autoencoder choice is based on multiple, possibly conflicting, criteria (accuracy and compression) to be optimized. There are many different existing algorithms that help in picking a best solution based on multiple criteria. In this study, the preferred method should be able to process any collection of solutions, as it is not known beforehand what kind of autoencoders will be involved. TOPSIS is a straight-forward and intuitive algorithm that can do this \cite{DBLP:journals/cor/HwangLL93}.

User-provided weights are applied to each of the criteria of each solution, allowing users to place emphasis on certain criteria. Then, the best and worst possible solutions, called the positive ideal alternative and the negative ideal alternative, are determined. In the context of convolutional autoencoders, the positive ideal has an accuracy and compression ratio of $1.0$, while the negative ideal has an accuracy and compression ratio of $0.0$. Finally, the ratio of $L^{2}$ distance between solutions and the ideal alternatives determines which solution is chosen.

The matter left open by this algorithm is that of the weights of criteria. Since there are no previous results to draw conclusion from, the weights will initially be set to equal values. The weights will depend on the relation between the autoencoder compression ratio and the accuracy of the classification networks.
\FloatBarrier

%% file: sections/4_experiments_results.tex
\section{Experiments and Results}

Experiments are performed on the popular CIFAR-10 data set \cite{krizhevsky2009learning}. It is one of the most popular data sets for the evaluation of CNNs, meaning that its use makes for easier comparisons with other approaches. It consists of $60000$ images, split into a training set of $50000$ images, and a test set of $10000$ images. $5000$ of the training set images are held out in a validation set, with the remaining $45000$ images constituting the actual training set. These data splits are randomized between experiments. Each image is a $32\text{x}32$ RGB image, and is associated with one of $10$ labels, which indicate the object in the image. The classes are evenly distributed over the training, validation, and test sets. In all experiments, stochastic gradient descent with a momentum of 0.9 is used to train networks. Training data is divided into batches of 50 samples, making for 1000 batches per epoch. Each network is trained for 25 epochs. This setup is copied from the baseline framework\cite{DBLP:conf/icml/RealMSSSTLK17} in order to help facilitate comparisons. 
Each worker process has access to an Nvidia Tesla K80 GPU. Concurrent access to the population of networks is facilitated purely by a shared file system; worker processes to not directly communicate with each other. By taking advantage of POSIX standards\footnote{\url{http://standards.ieee.org/develop/wg/POSIX.html}}, minimal parallel programming has been required \cite{breuel2010automlp}. The source code and accompanying documentation is available online.\footnote{\url{https://github.com/marijnvk/LargeScaleEvolution}} 

\subsection{Baseline Framework}

\begin{figure}[ht]
	\centering
	\includegraphics[width=1.2\textwidth]{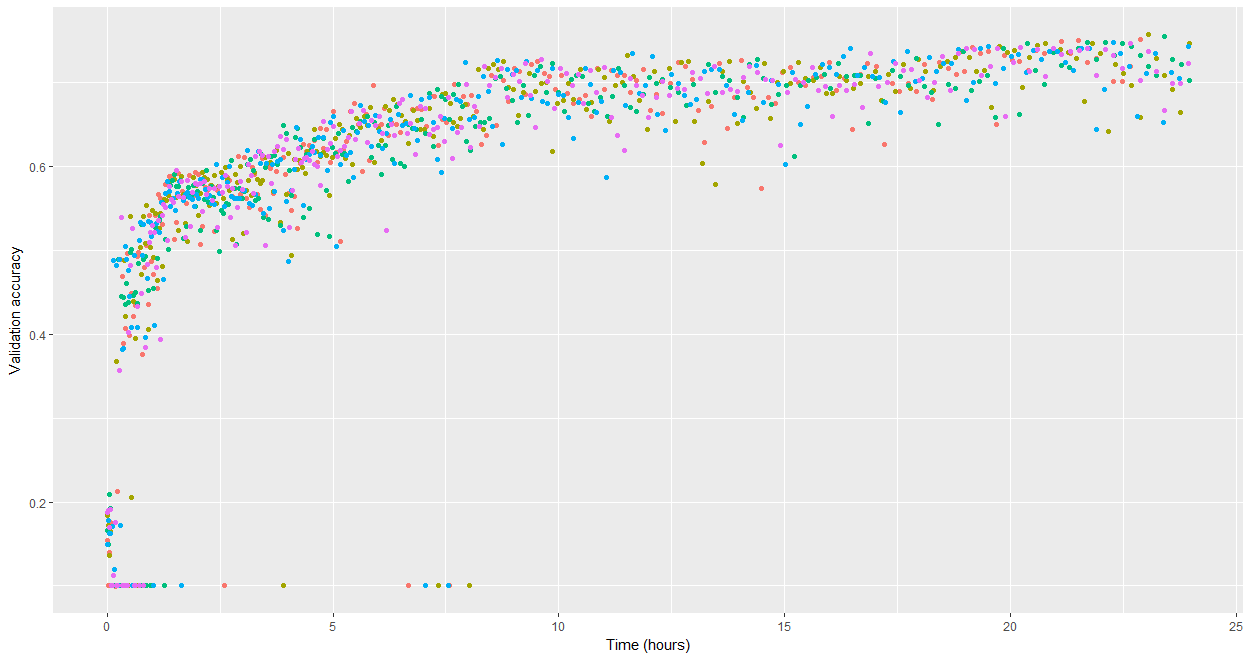}
	\caption{Results of the reproduction of the baseline framework, run with five worker processes. Each point indicates a completely trained CNN, color-coded by worker process.}
    \label{fig:run5_scatter}
\end{figure}

Initially, the baseline framework is run to confirm its performance, and to inspect its behavior in a more resource-restricted setting. The results of this experiment 
are listed in Table \ref{tab:original_compare}. Perhaps the most surprising result here is that after 12 hours, 5 worker processes have reached a much higher best classification accuracy than 250 worker processes. This may be partly because of the more limited available mutations, which ``focuses'' the evolutionary process better in the early stages in the case of five worker processes. At the 24 hour point, the baseline has best accuracy. Nonetheless, this shows that even a very small number of workers can be very effective at finding good neural networks. Figure \ref{fig:run5_scatter} illustrates the evolutionary process.

\begin{table}[ht]
	\caption{Performance results for the baseline setup of the framework. Accuracy values from the baseline study for the 12 and 24 hour interval are derived from a figure and thus are not precise. The baseline study did not report the number of generated networks.\\}
	\begin{tabular}[width=\textwidth]{ r | r | r | r | r}
    	\textbf{Study}&\textbf{\# Workers}&\textbf{Time elapsed (h)}&\textbf{Best accuracy (\%)}&\textbf{\# Networks}\\
        \hline
        \hline
        Baseline    &250& 12.0&$\sim$56.6&?\\
        Baseline    &250& 24.0&$\sim$86.9&?\\
        Baseline    &250&256.2&      94.6&?\\
        \hline
        This study& 10& 12.0&     72.32&848\\
        This study& 10& 24.0&     75.68&1197\\
        \hline
        This study&  1& 12.0&      58.7&190\\
        This study&  1& 24.0&      59.7&299\\
    \end{tabular}
    \label{tab:original_compare}
\end{table}

\subsection{Evolving Convolutional Autoencoders}

\begin{figure}[ht]
    \centering
    \begin{subfigure}[b]{0.47\textwidth}
        \includegraphics[width=\textwidth]{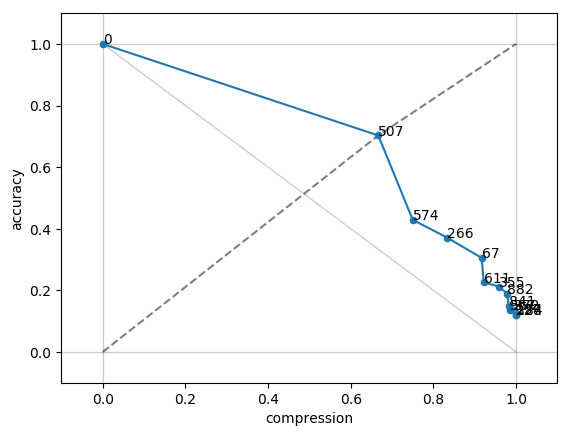}
        \caption{Pareto front}
    \end{subfigure}
    ~
    \begin{subfigure}[b]{0.47\textwidth}
        \includegraphics[width=\textwidth]{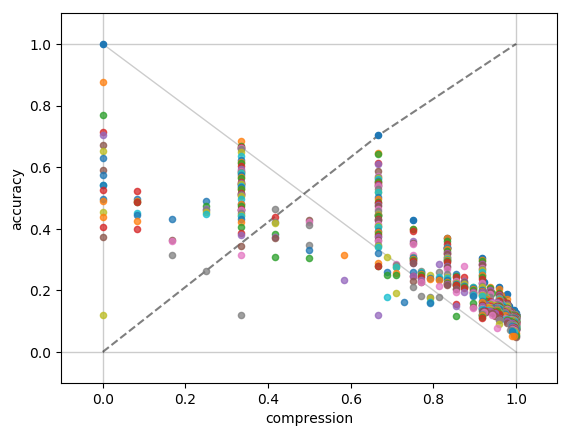}
        \caption{Full population}
    \end{subfigure}
    \caption{Results of the evolutionary process after generation 915. Each point illustrates a convolutional autoencoder, plotted by encoder compression and decoder accuracy. Fronts are marked by color. A compression of 1 means that samples are reduced to a single value. The individual chosen by TOPSIS is indicated by the dashed line. In the left plot, the number indicates the generation of that individual (i.e. the number of individuals that were generated before it).}
    \label{fig:p_950_results}
\end{figure}

The second experimental setting is the evolution of convolutional autoencoders. After evolving over 900 autoencoders in a span of two days on a single worker process, the status of the Pareto front is as in Figure \ref{fig:p_950_results}. The most telling structure in the population is the appearance of vertical groups of autoencoders. These groups share the same compression ratio. Generally those low in reconstruction accuracy tend to have fewer convolutional layers compared to those with a higher reconstruction accuracy. Their compression instead comes mostly from pooling layers. Evolution quickly pushes networks towards very high compression ratios, mostly due to the insertion of multiple pooling layers. Intuitively, however, it is the autoencoders that have a more even trade-off between compression ratio and reconstruction accuracy that are of most interest. These all have a fairly simple structure, usually being not more than five layers in depth. This is due to the nature of the training data. Images in the CIFAR-10 data set are already in compressed form when they are presented to the autoencoders. This undoubtedly reduces the effectiveness of the process step of evolving convolutional autoencoders.

At generation 507, an autoencoder is generated that clearly deviates from the otherwise quite linear shape of the final Pareto front. This individual attains a reconstruction accuracy of $70.3\%$ while reducing the input size by two-thirds. This autoencoder is expected to give the best results when we move to the last step of evolving image classifiers.

\subsection{Complete Framework}

The final setting combines both versions of the framework into a full approach that is supposed to outperform the baseline framework in terms of computation time at the cost of some accuracy. A number of representative autoencoders are picked from the Pareto font of evolved autoencoders and used to encode the CIFAR-10 data set. Table \ref{tab:cae_results} shows the chosen autoencoders and their characteristics.

Figure \ref{fig:cae_comp} shows a representative example of runs for each of the encoded data sets. As expected, the original data set results in the eventual highest classification accuracy. It is the data coming from autoencoder 507, the one which shows most promise and that progresses most similarly to the baseline. While the results are quite similar in terms of accuracy, the smaller, encoded data from autoencoder 507 means that approximately $20\%$ more networks are generated, trained, and evaluated. For more heavily compressed data sets, this increased number of generated networks is even higher, up to nearly $50\%$. As the number of generated and evaluated networks in the same time frame has increased, we have been able to increase the efficiency of the method. But on the other hand, in all other cases the encoding of data causes a significant loss of accuracy and early plateauing of evolutionary progress. It is clear that in the encoded setting, the evolutionary process has a significantly faster turn-over of networks. Note that this effect can be magnified if mutations are chosen in a more efficient way, which gives each new network a higher probability of being an improvement in terms of accuracy.

While the gains in this setting appear rather small, if applied to the baseline study, they would likely result in time savings on the scale of days. As the evolutionary process on classifiers is run longer, gains increase. After all, smaller sample sizes means that each batch is processed faster, and reduces the need for larger networks, which pushes down the average network size. Importantly, these results also show that the initial fair weighing of compression ratio and accuracy of the autoencoders for constructing the Pareto front is successful in determining the best autoencoder. Based on the results though, it may be advisable to weigh the autoencoder accuracy somewhat heavier since high compression rates seem to  cause heavy plateauing. Note that TOPSIS can easily be run again on the results of the evolution of autoencoders to obtain a differently-weighed result quickly.

\begin{table}[ht]
\centering
  \caption{Convolutional autoencoders (CAEs) sampled from the Pareto front. For each CAE, its generation ID, compression ratio, and mean classification accuracy over 5 runs are listed (variance in parentheses). Also listed is the average number of networks generated when evolving classifiers with training data encoded by this CAE. The last two CAEs do not come from the Pareto front, but share the same compression rate with the best CAE.\\}
  \begin{tabular}{ r | r | r | r }
    \textbf{CAE generation}&\textbf{Compression rate}&\textbf{Accuracy (Var)}&\textbf{\#Generated}\\
    \hline
    0& 1.0& 0.0 (0.0)&105.8\\
    507& 0.66& 0.7028 (0.0004)&121.6\\
    574& 0.75& 0.4289 (0.0004)&130.4\\
    266& 0.83& 0.3710 (0.0003)&136.9\\
    67& 0.91& 0.3054 (0.0003)&141.2\\
    611& 0.92& 0.2255 (0.0003)&140.9\\
    355& 0.95& 0.2123 (0.0003)&145.0\\
    882& 0.97& 0.1885 (0.0003)&144.2\\
    841& 0.98& 0.1493 (0.0002)&146.1\\
    \hline
    130&0.66&0.5650 (0.0004)&122.1\\
    725&0.66&0.4271 (0.0003)&118.4\\
  \end{tabular}
  \label{tab:cae_results}
\end{table}
\FloatBarrier

%% file: sections/5_conclusion.tex
\section{Conclusion}

A framework for evolving the structure of convolutional neural networks is introduced. Its main weakness is clearly the large amount of computational resources that are required to obtain networks with competitive accuracy. A method to remedy this weakness is proposed in this paper in the form of a two-step process. It consists of applying 
the baseline framework 
to the evolution of convolutional autoencoders in the hope of reducing the sample size of the input data. This makes any subsequent training of networks cheaper and helps in limiting their size.

A significant result is the viability of much smaller number of worker processes than previous studies demonstrated. This gives confidence in the other results of this study, and in future work with this framework. Experiments that investigate the impact of the new framework show the trade-off between compressing the input data and maintaining classification accuracy. The initial weighing of autoencoder accuracy and compression rate is deemed successful, while some tuning may help if the framework has to be used for other data sets. Although the experiments were necessarily limited in scope due to the available computational resources, the results indicate that this extended approach would result in significant running time reductions when applied at larger scales, both in terms of input size and computational time spent. 

\subsection{Future Work}
\label{sec:future}

The fundamental nature of the proposed work means that there are many more directions to explore. Mentioned here are some the more interesting ones.

\textbf{More complex data sets:}
As already mentioned before, the CIFAR-10 data set used in this study consists of images that are already compressed significantly. This naturally decreases the effectiveness of any convolutional autoencoder that attempts to reduce the image size further. Running the extended framework on a data set of larger images can realistically be expected to result in a much larger impact of the step that evolves the autoencoders.

\textbf{Beating humans at their own game:}
The initial population for the evolutionary process does not necessarily have to consist of very simple networks. An interesting application is to take state-of-the-art, human-designed networks as the individuals for the initial population. 

\textbf{Adaptive mutations:}
The availability of mutations affects the efficiency of the overall evolutionary process. Different kinds of mutations will be more or less likely to improve performance at different points in that process. The process would benefit from some sort of adaptive mutation choosing process, whereby mutations are sampled in a weighed fashion, perhaps even completely excluding some mutations at certain points in the process.

\textbf{Learning how to evolve:}
Given similar tasks, image classification for example, the evolutionary process can be expected to behave somewhat similarly in terms of which mutations are most effective at what points in the process. By adding a layer of abstraction on top of the evolutionary process which keeps track of this, it may be possible to learn how to best evolve networks given a task. 

\textbf{Constrained evolution for constrained networks:}
There are various situations, most notably in embedded systems, where compact networks are very desirable. The framework can easily be adapted to only consider a smaller solution space that can be limited in a variety of ways by placing additional constraints on the success of a mutation. 

\section{Acknowledgements}

Research leading to these results has received funding from the EU ECSEL Joint  
Undertaking under grant agreement no. 737459 (project  Productive4.0) and from Philips Research.